\documentclass[11pt]{article}

\usepackage[utf8]{inputenc}
\usepackage{deauthor}
\usepackage{times,graphicx}
\usepackage{amsfonts}
\usepackage{amsmath}
\usepackage{todonotes}
\usepackage{pifont}
\newcommand{\cmark}{\ding{51}}

\usepackage{multirow}
\usepackage{makecell}
\usepackage{booktabs}
\usepackage{caption,subcaption}
\usepackage{graphicx}

\title{Federated Learning without Full Labels: A Survey}

\author{Yilun Jin$^{\dagger}$
\hspace{2em} Yang Liu$^{\ddagger}$ 
\hspace{2em} Kai Chen$^{\dagger}$ 
\hspace{2em} Qiang Yang$^\dagger$ \\
$^{\dagger}$ Department of CSE, HKUST, Hong Kong, China \\
\texttt{\small yilun.jin@connect.ust.hk, \{qyang,kaichen\}@cse.ust.hk}\\
$^{\ddagger}$ Institute for AI Industry Research, Tsinghua University, Beijing, China \\ \texttt{\small liuy03@air.tsinghua.edu.cn}}

\usepackage{paralist}

\begin{document}

\maketitle
\begin{abstract}
Data privacy has become an increasingly important concern in real-world big data applications such as machine learning. To address the problem, federated learning (FL) has been a promising solution to building effective machine learning models from decentralized and private data. Existing federated learning algorithms mainly tackle the supervised learning problem, where data are assumed to be fully labeled. However, in practice, fully labeled data is often hard to obtain, as the participants may not have sufficient domain expertise, or they lack the motivation and tools to label data. Therefore, the problem of federated learning without full labels is important in real-world FL applications. In this paper, we discuss how the problem can be solved with machine learning techniques that leverage unlabeled data. We present a survey of methods that combine FL with semi-supervised learning, self-supervised learning, and transfer learning methods. We also summarize the datasets used to evaluate FL methods without full labels. Finally, we highlight future directions in the context of FL without full labels. 
\end{abstract}

\section{Introduction}
Deep learning (DL) algorithms have achieved great success in the past decade. Powered by large-scale data such as ImageNet \cite{deng2009imagenet}, ActivityNet \cite{heilbron2015activitynet}, BookCorpus \cite{zhu2015aligning}, and WikiText \cite{merity2017pointer}, deep learning models have been successfully applied to image classification \cite{he2016deep}, object detection \cite{he2017mask}, and natural language understanding \cite{devlin2019bert}. However, the success of DL relies on large-scale, high-quality data, which is not always available in practice for two reasons. On one hand, collecting and labeling data is costly, making it difficult for a single organization to accumulate and store large-scale data. On the other hand, it is also infeasible to share data across organizations to build large-scale datasets, as doing so leads to potential leakage of data privacy. In recent years, a series of laws and regulations have been enacted, such as the General Data Protection Regulation (GDPR) \cite{gdpr} and the California Consumer Privacy Act (CCPA) \cite{ccpa}, imposing constraints on data sharing. Therefore, how to jointly leverage the knowledge encoded in decentralized data while protecting data privacy becomes a critical problem. 

\textit{Federated Learning} (FL) \cite{yang2019federated,kairouz2021advances} is a promising solution to the problem and has received great attention from both the industry and the research community. The key idea of FL is that \textit{participants} (also known as \textit{clients} or \textit{parties}) exchange intermediate results, such as model parameters and gradients, instead of raw data, to jointly train machine learning models. As the raw data never leave their owners during model training, FL becomes an attractive privacy-preserving solution to the problem of decentralized machine learning. Up to now, a plethora of FL techniques has been proposed, focusing primarily on addressing the issues of data heterogeneity \cite{li2020federated,karimireddy2020scaffold}, system heterogeneity \cite{lai2021oort,diao2021heterofl}, data privacy and security \cite{bonawitz2017practical,zhang2020batchcrypt}, and communication efficiency \cite{mcmahan2017communication,reisizadeh2020fedpaq}. 

Despite the significant research efforts, there is still one important yet under-explored topic in FL, which is how to effectively leverage unlabeled data to learn better federated models. In existing efforts of FL \cite{mcmahan2017communication,li2020federated,lai2021oort}, it is assumed that all data held by all participants are fully labeled, and that a supervised learning problem is to be solved. However, the assumption may not hold in practice for two reasons. First, participants may not be sufficiently motivated to label their data. For example, suppose a sentiment classification model is to be trained with FL, smartphone users would be unwilling to spend time and effort to label all sentences typed in the phone. Second, participants may not have sufficient expertise to label their data. For example, wearable devices record various data (e.g. heart rate, breath rate, etc.) about the user's physical conditions, labeling which would require domain expertise in medical science and cannot be done by ordinary users. Based on the above observations, we argue that unlabeled data widely exist in real-world FL applications, and that the problem of \textit{Federated Learning without Full Labels} is an important problem to study. 

There are generally three learning paradigms in centralized machine learning (ML) that tackle the problem of learning without full labels, \textit{semi-supervised learning} \cite{zhu2005semi,chapelle2009semi}, \textit{self-supervised learning} \cite{liu2021self,jing2020self}, and \textit{transfer learning} \cite{pan2010survey}, all of which have drawn much attention from researchers. Among them, semi-supervised learning aims to leverage unlabeled data to assist the limited labeled data \cite{tarvainen2017mean,miyato2018virtual,berthelot2019mixmatch}. Self-supervised learning aims to learn indicative feature representations from unlabeled data, which are then used to assist downstream supervised learning tasks \cite{doersch2015unsupervised,gidaris2018unsupervised,he2022masked}. Transfer learning aims to use sufficient data from a \textit{source} domain to assist learning in a \textit{target} domain with insufficient data \cite{yosinski2014transferable,long2015learning,ganin2016domain}, where the target domain commonly contains unlabeled data. However, despite the large number of existing works in these areas, it is not straightforward to apply them in FL due to the following challenges. 
\begin{itemize}
    \item \textbf{Isolation of labeled and unlabeled data.} In traditional semi-supervised learning and transfer learning, the server has access to both labeled and unlabeled data. However, in FL without full labels, it is common for a participant to have unlabeled data only. For example, a medical institute may not have the expertise to diagnose a complex illness, leaving all its data unlabeled. Moreover, it is not allowed in FL to exchange labeled data to solve the problem. The isolation of labeled and unlabeled data may compromise the overall performance. As observed in \cite{diao2022semifl,jeong2021federated}, training with only unlabeled data leads to forgetting the knowledge learned from labeled data, which negatively impacts the overall performance. Therefore, it is important to bridge the knowledge between labeled and unlabeled data, without data exchange. 
    \item \textbf{Privacy of labeled data.} In the problem of FL without full labels, the number of labeled data is often limited. Therefore, participants have to repetitively access and exchange information about them to exploit the knowledge in the labels. This leads to risks of privacy leakage of the labeled data. For example, semi-honest participants can learn to reconstruct the labeled data via gradient inversion attacks \cite{geiping2020inverting}.
    \item \textbf{Data heterogeneity.} Data heterogeneity, i.e. the local data held by different participants have different data distributions, is an important property in FL that causes accuracy degradation \cite{li2020federated,lai2021oort}. Similarly, data heterogeneity also poses challenges in the problem of FL without full labels. For example, as the number of labeled data is limited, local models tend to overfit the local data more easily, which causes a greater amount of weight divergence \cite{zhao2018federated} and performance degradation. 
    \item \textbf{Balancing performance and efficiency.} The large-scale unlabeled data in the problem creates a tradeoff between performance and efficiency. Specifically, while large-scale unlabeled data is available for training, their impacts on the model performance may be marginal, and the overall efficiency can be improved by sampling a fraction of unlabeled data without compromising model performance. 
\end{itemize}

In this paper, we present a survey of the problem of FL without full labels and its existing solutions. The rest of the paper is organized as follows. Section \ref{sec:prelim} presents necessary backgrounds about FL as well as machine learning paradigms without full labels, including semi-supervised learning, self-supervised learning, and transfer learning. Sections \ref{sec:fed-semi}, \ref{sec:fed-self}, \ref{sec:fed-utl} then review methods on federated semi-supervised learning, federated self-supervised learning, and federated transfer learning, respectively. Section \ref{sec:dataset} summarizes the datasets used for evaluating FL methods without full labels. Section \ref{sec:related} analyzes the similarities and differences between our work and related surveys. Finally, Section \ref{sec:conclu-future} presents an outlook on potential directions in the context of FL without full labels.

\section{Preliminaries}\label{sec:prelim}
In this section, we formally introduce backgrounds about FL, as well as the machine learning paradigms leveraging unlabeled data, semi-supervised learning, self-supervised learning, and transfer learning. 
\subsection{Federated Learning (FL)}\label{sec:prelim-fl}
Federated Learning aims to virtually unify decentralized data held by different participants to train machine learning models while protecting data privacy. Depending on how the data is split across participants, FL can be divided into horizontal federated learning (HFL) and vertical federated learning (VFL) \cite{yang2019federated}. In HFL, participants own data with the same feature space (e.g. participants own image data from different users), while in VFL, participants own data with the same user space but different feature spaces (e.g. a financial institute owns transaction records of a user, while an e-commerce corporation owns purchase records). In this paper, following the majority of existing research efforts, we primarily focus on HFL\footnote{Unless otherwise specified, we will use FL to refer to HFL throughout this paper. }, i.e. all participants share the same feature space. Formally, we consider an FL scenario with $C$ participants, denoted as $1, \dots, C$. Each participant $i$ owns a dataset $\mathcal{D}_i=\{\mathbf{X}_{ij}, y_{ij}\}_{j=1}^{N_i}$, where $N_i=|\mathcal{D}_i|$ is the number of data held by participant $i$, and $\mathbf{X}_{ij}, y_{ij}$ denote the features and the label of the $j$-th sample from client $i$, respectively. We use $p_i(\mathbf{X}, y)$, $p_i(\mathbf{X})$, $p_i(y|\mathbf{X})$ to denote the joint distribution, marginal distribution, and conditional distribution of client $i$, respectively. Denoting the model parameters as $\theta\in \mathbb{R}^d$, the overall optimization objective of FL is as follows, 
\begin{equation}
    \min_{\theta} f_{fl}(\theta) = \frac{1}{C}\sum_{i=1}^C f_{fl, i}(\theta), \text{where } f_{fl, i}(\theta) =\frac{1}{N_i}\sum_{j=1}^{N_i}l\left(\mathbf{X}_{ij}, y_{ij};\theta\right), \text{s.t. } M_p(\theta)< \varepsilon_p,  
\label{eqn:fl}
\end{equation}
where $f_{fl, i}(\theta)$ is the local optimization objective of participant $i$, and $l(\mathbf{X}, y;\theta)$ is a loss function, such as the cross-entropy loss for classification problems. In addition, $M_p(\theta)$ denotes a metric measuring the privacy leakage of $\theta$ (e.g. the budget in differential privacy (DP) \cite{dwork2014algorithmic}), and $\varepsilon_p$ is a privacy constraint. 

The training process of FL generally involves multiple communication rounds, each of which contains two steps, local training, and server aggregation. 
\begin{itemize}
    \item In the local training stage, a subset of all participants is selected. They are given the latest global model and will train the model with their local data for several epochs. 

    \item In the server aggregation stage, participants upload their updated parameters to the server. The server aggregates received parameters via weighted averaging to obtain the global model for the next round. 
\end{itemize}

Depending on the properties of participants, FL can be categorized into \textit{cross-device} FL and \textit{cross-silo} FL \cite{kairouz2021advances}. Participants of cross-device FL are commonly smart devices (e.g. phones, sensors, wearables) connected with wireless networks, while participants of cross-silo FL are commonly large organizations with connected datacenters, implying the following differences: 
\begin{itemize}
    \item \textit{Computation/communication capability}. Participants in cross-device FL commonly have limited computation (e.g. small memory, limited power supply) and communication capability (e.g. wireless network). 
    \item \textit{Stability}. Participants in cross-device FL are not stable and may drop out due to network breakdown. 
    \item \textit{Participant states}. In general, participants in cross-device FL cannot carry state vectors, in that they may only participate in one round of FL, and then drop out indefinitely. 
\end{itemize}

\subsection{Machine Learning with Unlabeled Data}
\subsubsection{Semi-supervised Learning}\label{sec:prelim-semi}
In semi-supervised learning, there are two datasets, a labeled dataset $\mathcal{L} = \{\mathbf{X}_j, y_j\}_{j=1}^{|\mathcal{L}|}$, and an unlabeled dataset $\mathcal{U} = \{\mathbf{X}_k\}_{k=1}^{|\mathcal{U}|}$, $|\mathcal{U}|\ll |\mathcal{L}|$. In addition, the marginal distributions of $\mathcal{L}, \mathcal{U}$ are the same, i.e. $p_\mathcal{L}(\mathbf{X})=p_\mathcal{U}(\mathbf{X})$. The goal of semi-supervised learning, involving both labeled and unlabeled data, is as follows, 
\begin{equation}
\min_\theta f_{semi}(\theta) = \frac{1}{|\mathcal{L}|}\sum_{j=1}^{|\mathcal{L}|}l_s(\mathbf{X}_j, y_j;\theta)+\frac{1}{|\mathcal{U}|}\sum_{k=1}^{|\mathcal{U}|}l_u(\mathbf{X}_k;\theta),
\label{eqn:semi}
\end{equation}
where $l_s, l_u$ denotes the loss for labeled (supervised) and unlabeled data, respectively. 

We then introduce some widely adopted techniques in semi-supervised learning. 

\textbf{Pseudo-Labeling} \cite{lee2013pseudo}. Pseudo-labeling is a simple but effective trick for semi-supervised learning. Specifically, for each unlabeled data sample, its pseudo-label is taken as the class with the highest predicted probability, 
\begin{equation}
    \hat{y}_k = \arg\max_{c} g_\theta(\mathbf{X}_k)_c, 
\end{equation}
where $g_\theta$ is the model with parameter $\theta$, and $g_\theta(\mathbf{X}_k)_c$ denotes the predicted probability of class $c$ for $\mathbf{X}_k$. There is often a confidence threshold $\tau$, such that pseudo-labels are only taken on confident samples with $\hat{y}_k>\tau$. After that, the pseudo-labels are used to supervise learning on unlabeled data, i.e. 
\begin{equation}
    l_u(\mathbf{X}_k;\theta) = l_s(\mathbf{X}_k, \hat{y}_k;\theta).
\end{equation}

\textbf{Teacher-student Models} \cite{tarvainen2017mean}. Teacher-student models in semi-supervised learning leverage two networks, a teacher model $\theta_{tea}$ and a student model $\theta_{stu}$. On one hand, the student model is trained to be consistent with the teacher model to enhance its robustness 
\begin{equation}
    l_u(\mathbf{X}_k;\theta) = d\left(g_{\theta_{stu}}(\mathbf{X}_k), g_{\theta_{tea}}(\mathbf{X}_k)\right),
\end{equation}
where $d(\cdot, \cdot)$ is a distance metric. On the other hand, the teacher model is updated with moving averaging (parameterized by $\alpha$) over the student model after each iteration
\begin{equation}
\theta_{tea} =(1-\alpha)\theta_{tea} + \alpha \theta_{stu}.
\label{eqn:ema}
\end{equation}
\subsubsection{Self-supervised Learning}\label{sec:prelim-self}
\begin{table}[]
    \centering
    \resizebox{\linewidth}{!}{
    \begin{tabular}{ccccc}
    \toprule
        \multirowcell{2}{\textbf{Machine Learning Paradigm}} & \multicolumn{2}{c}{\textbf{Assumptions}} & \multirowcell{2}{\textbf{Application}} & \multirowcell{2}{\textbf{Limitations}} \\
        & Train-test i.i.d. & Labeled Data & \\
        \midrule
        Supervised Learning & \cmark & \cmark & Sufficient labeled data & Labeled data is hard to obtain. \\
        \midrule
        \multirowcell{2}{Semi-supervised Learning} & \multirowcell{2}{\cmark} & \multirowcell{2}{Insufficient} & A few labeled data &  \multirowcell{2}{Labeled and unlabeled data\\should have the same distribution. }\\
        & & & + large-scale unlabeled data & \\
        \midrule
        Self-supervised Learning & \cmark & $\times$ & Large-scale unlabeled data & Cannot directly perform supervised tasks. \\
        \midrule
        \multirowcell{3}{Transfer Learning} & \multirowcell{3}{$\times$} & \multirowcell{3}{From another domain} & Unlabeled data & \multirowcell{2}{Hard to select a helpful\\ source domain. \\Potential negative transfer. }\\
        & & & + labeled data \\
        & & & \textit{from another domain} \\
    \bottomrule
    \end{tabular}}
    \caption{A comparison between supervised learning, semi-supervised learning, self-supervised learning, and transfer learning.  }
    \label{tab:mlparadigm}
\end{table}

Self-supervised learning aims to learn good feature representations from unlabeled data to facilitate downstream machine learning tasks. There are in general two ways to perform self-supervised learning, generative learning, and contrastive learning \cite{liu2021self}. Generative learning trains the model to reconstruct the original data $\mathbf{X}$ from masked data to learn the internal semantics within $\mathbf{X}$, while contrastive learning trains the model to distinguish between `positive' and `negative' samples. In this survey, we primarily focus on contrastive learning, whose objective is given as follows. 
\begin{equation}
    \min_\theta f_{ctr}(\theta) = \sum_{\mathbf{X}\in\mathcal{U}}\left[d\left(g_\theta(\mathbf{X}), g_\theta(\mathbf{X}_+)\right) - \lambda\cdot d\left(g_\theta(\mathbf{X}), g_\theta(\mathbf{X}_{-})\right)\right],
    \label{eqn:self}
\end{equation}
where $\mathcal{U}$ is the unlabeled dataset, $g_\theta$ is a neural network parameterized by $\theta$, $\mathbf{X}_+, \mathbf{X}_-$ are positive and negative samples sampled for data $\mathbf{X}$, $\lambda$ is the weight for negative samples, and $d(\cdot, \cdot)$ is a distance metric. By minimizing $f_{ctr}$, the model $g_\theta$ learns to minimize the distance between positive samples in the feature space, while maximizing the distance between negative ones. Some representative contrastive learning methods include SimCLR \cite{chen2020simple}, MoCo \cite{he2020momentum}, BYOL \cite{grill2020bootstrap}, and SimSiam \cite{chen2021exploring}. We briefly explain their similarities and differences. 

\textbf{Similarities}. All four methods employ a Siamese structure -- two networks with the same architecture. One of them is called the online network $\theta_{o}$ and the other is called the target network $\theta_{tar}$. The main difference is that the online network is directly updated via gradient descent, while the target network is generally not. 

\textbf{Differences}. The differences between existing self-supervised learning methods are generally three-fold. 
\begin{enumerate}
    \item \textbf{Architecture}. In SimCLR and MoCo, the online and the target networks have the same architecture. On the contrary, for SimSiam and BYOL, the online network contains an additional predictor, i.e. $\theta_o = (\theta_o^f, \theta_o^p)$. The predictor aims to transform features between different views, enabling additional diversity. 
    \item \textbf{Target Network Parameter}. For SimCLR and SimSiam, the target network shares the same parameters as the online network $\theta_o=\theta_{tar}$, while for BYOL and MoCo, the target network is updated with an exponential moving average similar to Eqn. \ref{eqn:ema}.
    \item \textbf{Negative Samples}. On one hand, SimCLR and MoCo require negative samples $\mathbf{X}_-$. MoCo generates negative samples from previous batches, while SimCLR takes all other samples in the same batch as negative samples. On the other hand, SimSiam and BYOL do not require negative samples (i.e. $\lambda = 0$). 
\end{enumerate}

\subsubsection{Transfer Learning}\label{sec:prelim-unsup-trans}
Both semi-supervised learning and self-supervised learning assume that the training and test data are independent and identically distributed (i.i.d.), regardless of whether labels are present. However, transfer learning \cite{pan2010survey} does not require the assumption. Specifically, transfer learning deals with multiple data distributions (also called \textit{domains}) $p_i(\mathbf{X}, y), i=1, 2, \dots T$, where the model is trained on one, and tested on another. Without loss of generality, we assume that $T=2$. We denote $\mathcal{L}_1=\{\mathbf{X}_{1i}, y_{1i}\}_{i=1}^{|\mathcal{L}_1|}\sim p_1(\mathbf{X}, y)$ as the \textit{source} dataset, and $\mathcal{U}_2=\{\mathbf{X}_{2j}\}_{j=1}^{|\mathcal{U}_2|}\sim p_2(\mathbf{X})$ as the \textit{target} dataset. The overall goal is to minimize the error on the target dataset. However, as there are no labeled target data, we resort to the abundant source data to learn a model that generalizes well to the target dataset. A commonly studied optimization objective is as follows, 
\begin{equation}
    \min_{\theta_{f}, \theta_c}  \underbrace{\sum_{i=1}^{|\mathcal{L}_1|}l_s(\mathbf{X}_{1i}, y_{1i}; \theta_f, \theta_c)}_{f_{cls}(\mathcal{L}_1;\theta_f, \theta_c)} + \lambda\cdot \underbrace{d\left(g_{\theta_f}\left(\mathcal{L}_1\right), g_{\theta_f}\left(\mathcal{U}_2\right)\right)}_{f_{dom}({\mathcal{L}_1, \mathcal{U}_2};\theta_f)},
    \label{eqn:tl}
\end{equation}
where $\theta_f, \theta_c$, are parameters of the feature extractor and the classifier, respectively, $d(\cdot, \cdot)$ is a distance metric, $g_{\theta_f}(\mathcal{L}_1) = \{g_{\theta_f}(\mathbf{X}_{1i})\}_{i=1}^{|\mathcal{L}_1|}$ denotes the set of source features extracted by $\theta_f$, and $f_{cls}, f_{dom}$ denote the classifier loss on the source domain and the domain distance between domains, respectively. Intuitively, Eqn. \ref{eqn:tl} aims to minimize the classification error on the source domain, while minimizing the distance between source domain features and target domain features. In this way, the feature extractor $\theta_f$ is considered to extract domain-invariant features, and the classifier can be reused in the target domain. Commonly used distance metrics $d(\cdot, \cdot)$ include $L_2$ distance, maximum mean discrepancy (MMD) \cite{long2015learning} and adversarial domain discriminator \cite{ganin2016domain}. 

In addition, if an additional labeled target dataset $\mathcal{L}_2$ is available, $\theta_f, \theta_c$ can be further fine-tuned with $\mathcal{L}_2$. 

Transfer learning can generally be categorized into \textit{homogeneous transfer learning} and \textit{heterogeneous transfer learning} \cite{pan2010survey}. Homogeneous transfer learning assumes that domains share the same feature and label space, while heterogeneous transfer learning does not make such an assumption. For example, consider a movie recommender system that would like to borrow relevant knowledge from a book recommender system. If both systems rely on text reviews and ratings for recommendation, then a homogeneous transfer learning is to be solved, with the shared feature space being texts, and the shared label space being the ratings. However, if the movie recommender wants to leverage additional video clips, then the problem becomes a heterogeneous transfer learning problem, as the book recommender does not have video features. Heterogeneous transfer learning generally requires explicit cross-domain links to better bridge heterogeneous features and labels. For example, a novel and its related movie products should have similar feature representations. 
\subsubsection{Summary and Discussion}
We summarize the three learning paradigms involving unlabeled data in Table \ref{tab:mlparadigm}. As shown, supervised learning has two key assumptions, the i.i.d. property between training and test data, and sufficient labeled data. Therefore, supervised learning is not applicable when either the labeled data is insufficient, or the training and test data come from different distributions. To address the drawback, semi-supervised learning, self-supervised learning, and transfer learning are proposed to relax the two key assumptions. 
\begin{itemize}
    \item Semi-supervised learning relaxes the assumption of sufficient labeled data. With limited labeled data, semi-supervised learning aims to exploit large-scale unlabeled data that have the same distribution as labeled data with techniques such as pseudo-labeling or teacher-student models. The main limitation of semi-supervised learning is the difficulty to obtain i.i.d. unlabeled data. For example, for the task of medical imaging, the images taken from multiple hospitals may follow different distributions due to device differences, demographic shifts, etc. 
    \item Self-supervised learning further relaxes the assumption of labeled data. It aims to learn meaningful feature representations from the internal structures of unlabeled data, such as patches, rotations, and coloring in images. The main limitation of self-supervised learning is that, although it does not require labels to learn feature representations, they cannot be directly used to perform supervised tasks (e.g. classification).   
    \item Transfer learning further relaxes the assumption of i.i.d. train and test data. Given unlabeled data in a domain, it aims to learn from a different but related domain with sufficient labeled data, and to transfer helpful knowledge to the unlabeled data. The main limitation of transfer learning is that it commonly requires trial-and-errors to select an adequate source domain. When inadequate source domains are chosen, negative transfer \cite{rosenstein2005transfer} may happen which compromises model accuracy. 
\end{itemize}

\section{Federated Semi-supervised Learning}
\label{sec:fed-semi}
In this section, we present an overview of federated semi-supervised learning, whose main goal is to jointly use both labeled and unlabeled data owned by participants to improve FL. Before introducing detailed techniques, we first categorize federated semi-supervised learning into two settings following \cite{jeong2021federated}: 
\begin{itemize}
    \item \textbf{Label-at-client}, where the labeled data are located at the clients, while the server only has access to unlabeled data. For example, when a company would like to train an FL model for object detection using images taken from smartphones, the company has no access to the local data of users, and labeling can only be done by users. However, users are generally unwilling to label every picture taken from their smartphones, creating a label-at-client setting for federated semi-supervised learning. Formally, the objective function of this setting is as follows, 
    \begin{equation}
        %\begin{aligned}
        \min_{\theta} \frac{1}{C}\sum_{i=1}^Cf_{semi, i}(\theta), \text{s.t. } M_p(\theta)< \varepsilon_p\\
        %\text{where } f_i(\theta) &= \frac{1}{|\mathcal{L}_i|}\sum_{j=1}^{|\mathcal{L}_i|}l_s(\mathbf{X}_{ij}, y_{ij};\theta)+\frac{1}{|\mathcal{U}_i|}\sum_{k=1}^{|\mathcal{U}_i|}l_u(\mathbf{X}_{ik};\theta), ,
        %\end{aligned}
    \end{equation}
    where $f_{semi, i}(\theta)$ denotes the semi supervised learning loss (Eqn. \ref{eqn:semi}) evaluated on the dataset of participant $i$, and $M_p, \varepsilon_p$ follow Eqn. \ref{eqn:fl}. 
    \item \textbf{Label-at-server}, where the labeled data are located at the server, while clients have only unlabeled data. For example, consider a company of wearable devices that would like to train a health condition monitoring model with FL. In this case, users generally do not have the expertise to label data related to health conditions, leaving the data at clients unlabeled. The objective can be similarly formulated as 
    \begin{equation}
        \min_\theta \frac{1}{|\mathcal{L}|}\sum_{j=1}^{|\mathcal{L}|}l_s(\mathbf{X}_j, y_j;\theta) + \frac{1}{C}\sum_{i=1}^C\left(\frac{1}{|\mathcal{U}_i|}\sum_{k=1}^{|\mathcal{U}_i|}l_u(\mathbf{X}_{ik};\theta)\right), \text{s.t. }M_p(\theta)<\varepsilon_p. 
    \end{equation}
\end{itemize}

Methods for each federated semi-supervised learning setting are discussed in the following sections. We also summarize existing methods in Table \ref{tab:fed_ssl}. 

\begin{table}[t]
    \centering
    \resizebox{\linewidth}{!}{
    \begin{tabular}{cccccc}
    \toprule
        \multirowcell{2}{\textbf{Setting}} & \multirowcell{2}{\textbf{Method}} & \textbf{Label} & \textbf{Data} & \textbf{Data} & \textbf{Efficiency} \\
        & & \textbf{Isolation} & \textbf{Privacy} & \textbf{Heterogeneity} & \textbf{Tradeoff}\\
        \midrule
        \multirowcell{12}{Label-\\at-client} & \multirowcell{2}{RSCFed \cite{liang2022rscfed}} & Teacher-student & \multirowcell{2}{$\times$} & Sub-consensus models \&  & \multirowcell{2}{$\times$}\\
        & & model &  & distance-weighted aggregation & \\
        & \multirowcell{2}{FedSSL \cite{fan2022private}} & \multirowcell{2}{Pseudo-labeling} & \multirowcell{2}{Differential privacy (DP)} & \multirowcell{2}{Global generative model} & \multirowcell{2}{$\times$}\\
        & & \\
         & \multirowcell{2}{FedMatch \cite{jeong2021federated}} & \multirowcell{2}{Pseudo-labeling} & \multirowcell{2}{$\times$} & \multirowcell{2}{Inter-client consistency} & \multirowcell{2}{Disjoint \& sparse learning}\\
         & & & \\
         & \multirowcell{2}{FedPU \cite{lin2022federated}} & Negative labels  & \multirowcell{2}{$\times$} & \multirowcell{2}{$\times$} & \multirowcell{2}{$\times$}\\
         & & from other clients& & &\\
         & \multirowcell{2}{AdaFedSemi \cite{wang2022enhancing}} & \multirowcell{2}{Pseudo-labeling} & \multirowcell{2}{$\times$} & \multirowcell{2}{$\times$} & Tuning confidence threshold\\
         & & & & & and participation rate. \\
         & \multirowcell{2}{DS-FL \cite{itahara2021distillation}} & Ensemble & \multirowcell{2}{$\times$} & \multirowcell{2}{Entropy reduction averaging} & Transmit logits,\\
         & &pseudo-labeling & & & not parameters\\
         \midrule
        \multirowcell{4}{Label-\\at-server} & \multirowcell{2}{SemiFL \cite{diao2022semifl}}  & Alternate training \& & \multirowcell{2}{$\times$} & \multirowcell{2}{$\times$} & \multirowcell{2}{$\times$} \\
        & & Pseudo-labeling & \\
        & \multirowcell{2}{FedMatch \cite{jeong2021federated}} & Pseudo-labeling & \multirowcell{2}{$\times$} & Inter-client & \multirowcell{2}{Disjoint \& sparse learning}\\ 
        & & \& Disjoint learning & & consistency loss & \\ 
    \bottomrule
    \end{tabular}}
    \caption{Summary of techniques for federated semi-supervised learning. $\times$ indicates that the proposed method does not focus on this issue. }
    \label{tab:fed_ssl}
\end{table}

\subsection{The Label-at-client Setting}
The label-at-client setting of federated semi-supervised learning is similar to conventional FL (Eqn. \ref{eqn:fl}), in that clients can train local models with their labeled data, and the updated parameters are aggregated by the server. Therefore, the label-at-client setting inherits the challenges of data heterogeneity, data privacy, and efficiency tradeoff from conventional FL. In addition, some clients may not have labeled data to train their local models, causing the label isolation problem. We introduce how existing works address these problems in this section.

RSCFed \cite{liang2022rscfed} primarily focuses on the label isolation problem and the data heterogeneity problem in federated semi-supervised learning. For local training, the teacher-student model (introduced in Section \ref{sec:prelim-semi}) is adopted for training on unlabeled data. To further address the data heterogeneity problem, RSCFed proposes a sub-consensus sampling method and a distance-weighted aggregation method. In each round, several sub-consensus models are aggregated by independently sampling multiple subsets of all participants, such that each sub-consensus model is expected to contain participants with labeled data. Moreover, the local models are weighted according to their distance to sub-consensus models, such that deviating models receive low weights and their impacts are minimized.

FedSSL \cite{fan2022private} tackles the label isolation problem, the data privacy problem, and the data heterogeneity problem. To facilitate local training of unlabeled clients, FedSSL leverages the technique of pseudo-labeling. Further, to tackle the data heterogeneity problem, FedSSL learns a global generative model to generate data from a unified feature space, such that the data heterogeneity is mitigated by the generated data. Finally, to prevent privacy leakage caused by the generative model, FedSSL leverages differential privacy (DP) to limit the information leakage of the training data in the generative model.  

FedMatch \cite{jeong2021federated} proposes an inter-client consistency loss to address the data heterogeneity problem. Specifically, top-$k$ nearest clients are sampled for each client, and on each data sample, the output of the local model is regularized with those of the top-$k$ client models to ensure consistency. In addition, FedMatch proposes disjoint learning that splits the parameters for labeled and unlabeled data, and the parameters for unlabeled data are sparse. Upon updates, clients with only unlabeled data upload sparse tensors, reducing the communication cost. 

FedPU \cite{lin2022federated} studies a more challenging setting within semi-supervised learning, positive and unlabeled learning, in which each client has only labels in a subset of classes. In this setting, a client has only information about a part of all classes, leading to a severe label isolation problem. To tackle the problem, FedPU derives a novel objective function, such that the task of learning the negative classes of a client is relegated to other clients who have labeled data in the negative class. In this way, each client is only responsible for learning the positive classes and can do local training by itself. Empirically, the proposed FedPU outperforms FedMatch \cite{jeong2021federated} in the positive-and-unlabeled learning setting. 

AdaFedSemi \cite{wang2022enhancing} proposes a system to achieve the tradeoff between efficiency and model accuracy in federated semi-supervised learning with server-side unlabeled data. For every round, the model is trained with labeled data at clients and aggregated at the server. The server-side unlabeled data are incorporated into the training process via pseudo-labeling. AdaFedSemi \cite{wang2022enhancing} identifies two key parameters to balance the tradeoff between efficiency and performance, the client participation rate $P$, and the confidence threshold of pseudo-labels $\tau$. A lower $P$ reduces both the communication cost and the model accuracy, while a high $\tau$ reduces the server-side computation cost while also limiting the usage of unlabeled data. Therefore, AdaFedSemi designs a tuning method based on multi-armed bandits (MAB) to tune both parameters as training proceeds. Experiments show that AdaFedSemi achieves a good balance between efficiency and accuracy by dynamically adjusting $P$ and $\tau$ in different training phases. 

DS-FL \cite{itahara2021distillation} tackles a similar problem to AdaFedSemi, where clients own labeled data while the server owns unlabeled data. It proposes an ensemble pseudo-label solution to leverage the server-side unlabeled data. Specifically, instead of a single pseudo-label $\hat{y}_k$ for a data sample $\mathbf{X}_k$, it averages the pseudo-labels generated by all clients, i.e. $\hat{y}_k=\mathrm{MEAN}_{c=1}^C g_{\theta_c}(\mathbf{X}_k)$. This creates an ensemble of client models and offers better performance. Moreover, as only pseudo-labels are transmitted instead of model parameters, the communication cost can be significantly saved. In addition, DS-FL observes that training on pseudo-labels leads to a high prediction entropy. It then proposes an entropy-reduced aggregation, which sharpens the local outputs $g_{\theta_c}(\mathbf{X}_k)$ before aggregation. 

\subsection{The Label-at-server Setting}
The label-at-server setting, where clients do not have any labeled data, is more challenging than the label-at-client setting. The reason is that all clients own unlabeled data only and cannot provide additional supervision signals to the FL model. As shown in \cite{jeong2021federated} and \cite{diao2022semifl}, training with only unlabeled data may lead to catastrophic forgetting of the knowledge learned from labeled data, and thus compromises the model performance. 

To address the isolation between labeled data and unlabeled data, FedMatch \cite{jeong2021federated} proposes a disjoint learning scheme that involves two sets of parameters for labeled and unlabeled data, respectively. The parameters for labeled data are fixed when training on unlabeled data, and vice versa, to prevent the knowledge from being overwritten. Disjoint learning brings additional benefits in communication efficiency, in that the parameters for unlabeled data, which are transmitted between participants and the server, are set to be sparse. In addition, to address the heterogeneous data held by different clients, FedMatch proposes an inter-client consistency loss, such that local models from different participants generate similar outputs on the same data. 

SemiFL \cite{diao2022semifl} takes another approach to solving the challenges. It proposes to fine-tune the global model with labeled data to enhance its quality and to alleviate the forgetting caused by unsupervised training at clients. Furthermore, instead of regularizing model outputs across clients, SemiFL proposes to maximize the consistency between client models and the global model. Specifically, the global model generates pseudo-labels for client-side unlabeled data, and the local models of clients are trained to fit the pseudo-labels. Empirical results show that SemiFL yields more competitive results than FedMatch.

\section{Federated Self-supervised Learning}\label{sec:fed-self}
In this section, we introduce how self-supervised learning can be combined with FL to learn with decentralized and purely unlabeled data. Although there are two types of self-supervised learning, generative and contrastive learning, so far only contrastive methods have been studied in the FL setting, and thus we limit the discussions within federated contrastive self-supervised learning. The objective function can be formalized as 
\begin{equation}
    \min_\theta \frac{1}{C}\sum_{i=1}^Cf_{ctr, i}(\theta), \text{s.t. }M_p(\theta)< \varepsilon_p,
\end{equation}
where $f_{ctr, i}$ denotes $f_{ctr}$ (Eqn. \ref{eqn:self}) evaluated at participant $i$. Compared to FL with full supervision, federated contrastive learning does not have globally consistent labels, and thus, the local contrastive objectives may deviate from one another to a greater extent. Therefore, heterogeneous data poses a greater challenge to federated contrastive learning. Table \ref{tab:fed_self} summarizes existing works in federated contrastive self-supervised learning. 
\begin{table}[t]
    \centering
    \resizebox{\linewidth}{!}{
    \begin{tabular}{ccccc}
    \toprule
        \multirowcell{2}{\textbf{Method}} & \textbf{Label} & \textbf{Data} & \textbf{Data} & \textbf{Efficiency} \\
        & \textbf{Isolation} & \textbf{Privacy} & \textbf{Heterogeneity} & \textbf{Tradeoff}\\
        \midrule
        FedCA \cite{zhang2020federated} & SimCLR & $\times$ & Dictionary \& Alignment module & $\times$\\
        SSFL \cite{he2021ssfl} & SimSiam & $\times$ & Personalized models & $\times$\\
        FedU \cite{zhuang2021collaborative} & BYOL & $\times$ & Selective divergence-aware update & $\times$\\
        FedEMA \cite{zhuang2022divergence} & BYOL & $\times$ & Moving average client update & $\times$\\
        FedX \cite{han2022fedx} & Local relation loss & $\times$ & Global contrastive \& relation loss & $\times$\\ 
        \multirowcell{2}{Orchestra \cite{lubana2022orchestra}} & Rotation prediction & Sending local centroids & \multirowcell{2}{Bi-level clustering} & \multirowcell{2}{$\times$}\\
        & \& clustering & instead of all representations  & & \\
    \bottomrule
    \end{tabular}}
    \caption{Summary of techniques for federated self-supervised learning. $\times$ indicates that the proposed method does not focus on this issue. }
    \label{tab:fed_self}
\end{table}

FedCA \cite{zhang2020federated}, as one of the earliest works to study federated self-supervised learning, proposes a dictionary module and an alignment module to solve the feature misalignment problem caused by data heterogeneity. Extending SimCLR, the dictionary module in FedCA aims to use the global model to generate consistent negative samples across clients, while the alignment module uses a set of public data to align the representations generated by local models. However, the alignment module of FedCA requires sharing a public dataset, which compromises data privacy. 

SSFL \cite{he2021ssfl} addresses the data heterogeneity problem in federated self-supervised learning with a personalized FL framework \cite{tan2022towards,li2021ditto}, in which each participant trains s unique local model instead of training a shared global model. The drawback of SSFL is that the adopted self-supervised learning method requires a large batch size, which is hard to achieve on resource-limited edge devices. 

FedU \cite{zhuang2021collaborative} designs a heterogeneity-aware aggregation scheme to address data heterogeneity in federated self-supervised learning. As discussed in Section \ref{sec:prelim-self}, there are generally two networks in contrastive learning, an online network and a target network. Therefore, how to aggregate and update the two networks in FL with data heterogeneity becomes an important research question. With empirical experiments, FedU discovers that aggregating and updating only the online network yields better performances. Moreover, as FedU extends BYOL with an additional predictor model, it is also necessary to design an update rule for it. FedU designs a divergence-aware predictor update rule, which updates the local predictor only when its deviation from the global predictor is low. These rules ensure that data heterogeneity is well captured by local and global models.  

Extending FedU, FedEMA \cite{zhuang2022divergence} presents an extensive empirical study on the design components of federated contrastive learning. It performs experiments combining FL with MoCo, SimCLR, SimSiam, and BYOL, and identifies BYOL as the best base method. Based on the results, FedRMA with a divergence-aware moving average update rule is proposed. The difference between FedEMA and FedU is that, FedU overwrites the local online model with the global online model 
\begin{equation}
    \theta_{o, c}^{r} = \theta_{o}^{r-1},
\end{equation}
where $\theta_{o, c}^r$ denotes the local online model at client $c$ and round $r$, and $\theta_o^{r-1}$ denotes the global online model aggregated at the previous round. On the contrary, FedEMA updates the local online model by interpolating between the global and local online models to adaptively incorporate global knowledge, i.e. 
\begin{equation}
    \theta_{o, c}^r = (1-\mu)\theta_{o}^{r-1} + \mu \theta_{o,c}^{r-1}, 
\end{equation}
where $\mu$ is a parameter based on weight divergence, 
\begin{equation}
    \mu = \min (\lambda\|\theta_{o, c}^{r-1}-\theta_o^{r-1}\|, 1).
\end{equation}
While FedU and FedEMA are simple and effective, they both require \textit{stateful} clients to keep track of the divergence between local and global models, and are thus not applicable in cross-device FL. 

Contrary to FedU and FedEMA, Orchestra \cite{lubana2022orchestra} proposes a theoretically guided federated self-supervised learning method that works with cross-device FL. Orchestra is based on the theory that feature representations with good clustering properties yield low classification errors. Therefore, in addition to contrastive learning, Orchestra aims to simultaneously enhance the clustering properties of all data representations. However, sharing all data representations for clustering may cause the problem of privacy leakage. Orchestra addresses the problem with a bi-level clustering method, in which clients first cluster their data representations, and send only the local centroids to the server. The server performs a second clustering on the local centroids to obtain global clustering centroids, which are sent back to clients to compute cluster assignments. As local centroids reveal less information than all data representations, this bi-level clustering method better preserves data privacy. 

Orthogonal to the above methods, FedX \cite{han2022fedx} proposes a versatile add-on module for federated self-supervised learning methods. FedX consists of both local and global relational loss terms that can be added to various contrastive learning modules. The local relational loss aims to ensure that under the local model, two augmentations of the same data sample have similar relations (similarities) to samples within the same batch $\mathcal{B}$, i.e. 
\begin{equation}
        \mathbf{r}_i^j = \frac{\exp(\mathrm{sim}(\mathbf{z}_i, \mathbf{z}_j))}{\sum_{k\in\mathcal{B}}\exp(\mathrm{sim}(\mathbf{z}_i, \mathbf{z}_k))}, \mathbf{r}_{i+}^{j} = \frac{\exp(\mathrm{sim}(\mathbf{z}_{i+}, \mathbf{z}_j))}{\sum_{k\in\mathcal{B}}\exp(\mathrm{sim}(\mathbf{z}_{i+}, \mathbf{z}_k))}, 
\end{equation}
\begin{equation}
        L_{rel} = \mathrm{JS}(\mathbf{r}_i, \mathbf{r}_{i+}),
\end{equation}
where $\mathbf{z}_i, \mathbf{z}_{i+}$ denotes the feature representation of the $i$-th sample (with different augmentations) in the batch, $\mathbf{r}_i$ denotes the (normalized) similarity between the $i$-th sample and all other samples, and $\mathrm{JS}$ denotes the Jensen-Shannon divergence. The global relational loss is similarly defined such that under the local model, two augmentations of the same data sample have similar relations to the global representations. Empirical results show that FedX is versatile and can improve the performance of various contrastive learning methods in FL, including FedSimCLR, FedMoCo, and FedU.

\section{Federated Transfer Learning}
\label{sec:fed-utl}
In this section, we summarize efforts that combine FL with transfer learning (FTL). We categorize existing works in FTL into homogeneous FTL and heterogeneous FTL, whose differences are introduced in Section \ref{sec:prelim-unsup-trans}. 

\subsection{Homogeneous FTL}
In this section, we introduce research works on homogeneous FTL. Assuming that there are $S$ source domains and $T$ target domains, each of which is held by one participant, the objective of homogeneous FTL is as follows,
\begin{equation}
\min_{\theta_f, \theta_c} \sum_{i=1}^Sf_{cls}(\mathcal{L}_i;\theta_f, \theta_c) + \sum_{i=1}^S\sum_{j=1}^T\lambda_{ij}f_{dom}(\mathcal{L}_i, \mathcal{U}_j;\theta_f), \text{s.t. } M_p(\theta_f, \theta_c) < \varepsilon_p, 
\end{equation}
where $\mathcal{L}_i, \mathcal{U}_j$ denote the labeled/unlabeled dataset held by the $i$-th source/target domain, respectively, $f_{cls}, f_{dom}$ follow Eqn. \ref{eqn:tl}, and $\lambda_{ij}$ are hyperparameters used to select helpful source domains. If source domain $i$ and target domain $j$ are similar, we can assign a high $\lambda_{ij}$, and vice versa. Depending on how many source or target domains are involved, we can categorize existing works into two settings, single-source, and multi-source. In the multi-source setting, selecting the most appropriate source domain poses an additional challenge compared to the single-source setting. We introduce related works in both settings in the following sections. 

\subsubsection{Single-source Setting}
The single-source setting in federated transfer learning commonly involves one server with labeled data, and multiple clients with unlabeled data. As the clients themselves may have different data distributions, each client creates a unique target domain, which requires a flexible adaptation method to tackle multiple targets.  

To our knowledge, DualAdapt \cite{yao2022federated} is the first work to tackle the single-source, multi-target federated transfer learning problem. DualAdapt extends from the maximum classifier discrepancy (MCD) method \cite{saito2018maximum}. Specifically, MCD involves a feature extractor $\theta_f$ and two classifiers $\theta_{c1}, \theta_{c2}$, trained with the following steps iteratively: 
\begin{itemize}
    \item First, $\theta_f, \theta_{c1}, \theta_{c2}$ are trained to minimize the error on the source domain $\mathcal{L}_1$. 
    \item Second, given a target domain sample $\mathbf{X}_t$, we fix the feature extractor $\theta_f$ and maximize the discrepancy between the classifiers, i.e. $\max_{\theta_{c1}, \theta_{c2}}L_{cd} = d(g_{\theta_f, \theta_{c1}}(\mathbf{X}_t), g_{\theta_f, \theta_{c2}}(\mathbf{X}_t))$. This step aims to find target samples that are dissimilar to the source domain. 
    \item Third, the classifiers are fixed, and the feature extractor $\theta_f$ is trained to minimize $L_{cd}$ to generate domain invariant features. 
\end{itemize}
The FL setting creates two challenges for MCD. First, Step 2 should be taken at clients, yet as no labels are available, Step 2 may result in naive non-discriminative solutions. To address the problem, DualAdapt proposes client self-training, where pseudo-labels generated by the server model are used to train the classifiers in addition to $L_{cd}$. Second, to maintain a single feature extractor $\theta_f$, Step 3 is done at the server, which has no access to target samples $\mathbf{X}_t$. DualAdapt proposes to use mixup \cite{zhang2017mixup} to approximate target samples $\mathbf{X}_t$. To further mitigate the impact of domain discrepancy, DualAdapt proposes to fit Gaussian mixture models (GMM) at each participant. At each participant, samples from other participants are re-weighed via the fit GMMs, such that impacts of highly dissimilar samples are mitigated.

FRuDA \cite{gan2022fruda} proposes a system for single-source, multi-target federated transfer learning with DANN \cite{ganin2016domain} Similar to DualAdapt, it also considers the setting with multiple unlabeled target domains, for which it proposes an optimal collaboration selection (OCS) method. The intuition of OCS is that, for a new target domain, instead of always transferring from the only source domain, it is also possible to transfer from an existing target domain that is closer to the new domain. To implement the intuition, OCS derives an upper bound for the transfer learning error from one domain to another, 
\begin{equation}
    \varepsilon_{CE, D_2}(h, h') \le \theta_{CE}(\varepsilon_{L_1, D_1}(h, h') + 2\theta W(D_1, D_2)),
    \label{eqn:ocs}
\end{equation}
where $\varepsilon_{M, D}(h, l)$ denotes the error, measured by metric $M$, of the hypothesis $h$ on data distribution $D$ with the label function $l$, $\theta_{CE}, \theta$ are constants, $h, h'$ are source and target hypotheses, $D_1, D_2$ are source and target data distributions, respectively, and $W(D_1, D_2)$ denotes the Wasserstein distance between $D_1, D_2$. With Eqn. \ref{eqn:ocs}, the optimal collaborator of each target domain can be selected by minimizing the right-hand side. To further improve efficiency, a lazy update scheme, exchanging discriminator gradients every $p$ iteration, is further proposed. 

\begin{table}[t]
    \centering
    \resizebox{\linewidth}{!}{
    \begin{tabular}{cccccccc}
    \toprule
        \multirowcell{2}{\textbf{Method}} & \multirowcell{2}{\textbf{Homo}-\\ \textbf{geneous}} & \multirowcell{2}{\# \textbf{Source}} & \multirowcell{2}{\# \textbf{Target}} & \textbf{Label} & \textbf{Data} & \textbf{Data} & \textbf{Efficiency} \\
        & & & & \textbf{Isolation} & \textbf{Privacy} & \textbf{Heterogeneity} & \textbf{Tradeoff}\\
        \midrule
        \multirowcell{2}{DualAdapt \cite{yao2022federated}} & \multirowcell{2}{\cmark} & \multirowcell{2}{1} & \multirowcell{2}{$>1$} & MCD \& Pseudo-labeling\&  & \multirowcell{2}{$\times$} & \multirowcell{2}{GMM weighting} & \multirowcell{2}{$\times$}  \\
        & & & & MixUp approximation\\
        \multirowcell{2}{FRuDA \cite{gan2022fruda}} &\multirowcell{2}{\cmark} & \multirowcell{2}{1} & \multirowcell{2}{$>1$} & DANN \cite{ganin2016domain} \& & \multirowcell{2}{$\times$} & \multirowcell{2}{Optimal collaborator selection} & \multirowcell{2}{Lazy update}\\
        & & & & Optimal collaborator selection\\
        \multirowcell{2}{FADA \cite{peng2020federated}} & \multirowcell{2}{\cmark} & \multirowcell{2}{$>1$} & \multirowcell{2}{1} & DANN \cite{ganin2016domain} \& & \multirowcell{2}{$\times$} & \multirowcell{2}{Gap statistics weighting} & \multirowcell{2}{$\times$}\\
        & & & & Representation sharing\\
        \multirowcell{2}{FADE \cite{hong2021federated}} & \multirowcell{2}{\cmark} & \multirowcell{2}{$>1$} & \multirowcell{2}{1} & \multirowcell{2}{DANN}  & \multirowcell{2}{No representation sharing} & CDAN  & \multirowcell{2}{$\times$}\\
        & & & && & Squared adversarial loss & \\
        EfficientFDA \cite{kang2022communicational} & \cmark & $>1$ & 1 & Max. mean discrepancy (MMD) & Homomorphic encryption (HE) & $\times$ & Optimized HE operation\\
        PrADA \cite{kang2022privacy} & \cmark & 2 & 1 & Grouped DANN & Homomorphic encryption (HE) & $\times$ & $\times$\\
        SFTL \cite{liu2020secure} & $\times$ & 1 & 1 & Sample alignment loss & HE \& Secret sharing (SS) & Sample alignment loss & $\times$\\
        SFHTL \cite{feng2022semi} & $\times$ & 1 & $>$1 & Label propagation & Split learning \cite{vepakomma2018split} & Unified feature space & $\times$ \\

    \bottomrule
    \end{tabular}}
    \caption{Summary of techniques for (unsupervised) federated transfer learning. $\times$ indicates that the proposed method does not focus on this issue. 'Homogeneous' indicates whether the work focuses on homogeneous FTL (\cmark) or heterogeneous FTL ($\times$). \# Source, \# Target denote the number of source and target domains considered in the work, respectively. }
    \label{tab:fed_utl}
\end{table}

\subsubsection{Multi-source Setting}
A more challenging setting of federated transfer learning is the multi-source setting, where multiple source domains with labeled data are available to transfer knowledge to a single unlabeled target domain. In this setting, it is necessary to select a source domain with helpful knowledge without directly observing source data. 

To our knowledge, FADA \cite{peng2020federated} is the first work to tackle the multi-source federated transfer learning problem. FADA extends the adversarial domain adaptation \cite{ganin2016domain} method, with a domain discriminator between each source domain and the target domain. The domain discriminator aims to tell whether each feature representation belongs to the source and the target domain, and the feature extractor is then trained to fool the domain discriminator to learn domain invariant features. To train the domain discriminator, FADA directly exchanges feature representations from both domains, which may lead to potential privacy threats. In addition, to select the most relevant source domain to transfer from, FADA proposes a source domain weighting method based on gap statistics. Gap statistics \cite{tibshirani2001estimating} measures how well the feature representations are clustered, 
\begin{equation}
    I = \sum_{r=1}^k\frac{1}{2n_r}\sum_{i, j\in C_r}\|\mathbf{z}_i-\mathbf{z}_j\|_2,
\end{equation}
where $\mathbf{f}_i$ denotes the feature representation of the $i$-th sample, $C_1\dots C_k$ denote the index set of $k$ clusters, and $n_r$ is the number of samples in cluster $r$. A low $I$ indicates that the feature representations can be clustered with low intra-cluster variance, which usually indicates good features. FADA then computes how the gap statistics of the target domain drop after learning with each source domain, i.e. 
\begin{equation}
    I_i^{gain} = I_i^{r-1}-I_i^{r}, 
\end{equation}
where $r$ denotes the communication round, and $i$ denotes the source domain index. Finally, FADA applies weights on source domains via with $\mathrm{Softmax}(I_1^{gain}, I_2^{gain}\dots, )$. 

FADE \cite{hong2021federated} improves over FADA by not sharing representations to learn the domain discriminator, thus better protecting data privacy. Instead, the domain discriminator is kept local at each client, and is trained locally and updated via parameter aggregation. FADE theoretically shows that the design leads to the same optimal values as FADA, but empirically leads to negative impacts. The issues of the design are that the trained discriminator may have low sensitivity (and thus takes longer to converge) and user mode collapse (and thus fail to represent heterogeneous data). To address the drawbacks, FADA presents two tricks. To tackle the low sensitivity issue, FADE squares the adversarial loss such that it is more reactive under large loss values. To tackle the user mode collapse issue, FADE proposes to maximize the mutual information between users (related to classes) and representations, and implements the idea with conditional adversarial domain adaptation (CDAN) \cite{long2018conditional}.  

EfficientFDA \cite{kang2022communicational} is another improvement over FADA in that source and target domain feature representations are encrypted with homomorphic encryption (HE) \cite{aono2017privacy}, and the maximum mean discrepancy (MMD) \cite{long2015learning} is computed over ciphertexts. As homomorphic encryption incurs large computation and communication costs, EfficientFDA further proposes two ciphertext optimizations. First, ciphertexts in each batch of samples are aggregated to reduce communication overhead. Second, for computing gradients with ciphertexts, the chain rule is applied to replace ciphertext computations with plaintexts to improve computational efficiency. Experiments show that EfficientFDA achieves privacy in federated transfer learning, while being 10-100x more efficient than naive HE-based implementations.  

While the above works tackle the problem with multiple source domains with the same feature space, PrADA \cite{kang2022privacy} tackles a different problem, involving two source domains with different feature spaces. PrADA considers a partially labeled target domain A $\{\mathbf{X}_l^A, y_l^A\}\cup \{\mathbf{X}^A_u\}$, a labeled source domain B $\{\mathbf{X}^B\in\mathbb{R}^{N_B\times D}, y^B\}$, and a feature source domain C $\{\mathbf{X}^A_C\in \mathbb{R}^{N_A\times D_C}\}\cup \{\mathbf{X}^B_C\in\mathbb{R}^{N_B\times D_C}\}$. Domains A and B share the same feature space with different distributions, while domain C aims to provide rich auxiliary features for samples in both A and B. PrADA presents a fine-grained domain adaptation technique, in which features from domain C are first manually grouped into $g$ tightly relevant feature groups. Each feature group is then assigned a feature extractor and a domain discriminator to perform fine-grained, group-level domain adaptation. In addition, to protect data privacy, the whole training process is protected with homomorphic encryption. Experiments show that with the grouped domain adaptation, PrADA achieves better transferability and interpretability. 

\subsection{Heterogeneous FTL}
In this section, we introduce existing works about heterogeneous FTL. Compared to homogeneous FTL, the main difference of heterogeneous FTL is that it commonly requires cross-domain links between data (e.g. different features of the same user ID, the same features from different users, etc.) to bridge the heterogeneous feature spaces. Formally, assuming a heterogeneous FTL setting with two parties, A and B, with data $\mathcal{D}_A, \mathcal{D}_B$, with $\mathcal{D}_{AB}=\mathcal{D}_A\cap \mathcal{D}_B$ being the overlapping dataset (i.e. cross-domain links), the objective of heterogeneous FTL is 
\begin{equation}
    \min_{\theta_A, \theta_B} L_{A}(\mathcal{D}_A;\theta_A) + L_B(\mathcal{D}_B;\theta_B) + \lambda L_{algn}(\mathcal{D}_{AB};\theta_A, \theta_B), \text{s.t. } M_p(\theta_A, \theta_B)< \varepsilon_p,
\end{equation}
where $L_A, L_B$ are loss functions on dataset $\mathcal{D}_A, \mathcal{D}_B$, respectively, and $L_{algn}$ is an alignment loss that aims to align the overlapping dataset $\mathcal{D}_{AB}$ between domains. However, in FL, sharing sample features or labels pose potential privacy threats. How to leverage the cross-domain sample links to transfer knowledge while preserving privacy thus becomes a key challenge to solve. 

To our knowledge, SFTL \cite{liu2020secure} is the first work to tackle the heterogeneous FTL problem. It considers a two-party setting and assumes that some user IDs $\mathcal{I}_{AB}$ exist in both parties (with different features). SFTL proposes an alignment loss to minimize the difference between features of the same users to achieve knowledge transfer, 
\begin{equation}
    L_{algn} = \sum_{i\in\mathcal{I}_{AB}}d(g_{\theta_A}(\mathbf{X}_i^A), g_{\theta_B}(\mathbf{X}_i^B)),
\end{equation}
where $\mathcal{I}_{AB}$ denotes the overlapping user ID set, $g_{\theta_A}, g_{\theta_B}$ denote neural network models of party A and B, and $\mathbf{X}_i^A, \mathbf{X}_i^B$ denote the features of user $i$ held by party A and B, respectively. In addition, SFTL addresses the data privacy problem by designing two secure protocols for SFTL, one based on homomorphic encryption, and the other based on secret sharing (SS). 

The drawbacks of SFTL are that it is limited to the two-party setting, and both A and B have only partial models and cannot perform independent inference. To address these drawbacks, SFHTL \cite{feng2022semi} proposes an improved framework that supports multiple parties. The main difficulty in the multi-party heterogeneous FTL is the lack of overlapping samples and labels. To address the lack of overlapping samples, SFHTL proposes a feature reconstruction technique to complement the missing non-overlapping features. Specifically, all parties are trained to project their features into a unified latent feature space. Then, each party learns a reconstruction function that projects the unified features to raw features. With the reconstruction functions, each party can expand the feature spaces of non-overlapping samples, thus enlarging the training dataset. In addition, SFHTL proposes a pseudo-labeling method based on label propagation \cite{zhu2005semi} to address the lack of labels. Specifically, a nearest neighbor graph based on feature proximity in the unified feature space is constructed, and the labels are propagated from labeled samples to unlabeled samples via the graph. Finally, to protect the privacy of labels, SFHTL is trained with split learning, such that labels are not directly shared with other parties. 

\section{Datasets and Evaluations}\label{sec:dataset}
\begin{table}[]
    \centering
    \resizebox{\linewidth}{!}{
    \begin{tabular}{cccccccc}
    \toprule
        \multirowcell{2}{\textbf{Dataset}} & \multicolumn{3}{c}{\textbf{FL Methods without Full Labels}} & \multirowcell{2}{\textbf{Application}} & \multirowcell{2}{\# \textbf{Domains}} & \multirowcell{2}{\# \textbf{Samples}} & \multirowcell{2}{\textbf{Partition}}\\
        & \textbf{Semi} & \textbf{Self} & \textbf{Trans.} & \\
    \midrule
        CIFAR-10& \cmark \cite{liang2022rscfed,fan2022private,jeong2021federated,wang2022enhancing,diao2022semifl} & \cmark \cite{he2021ssfl,zhuang2021collaborative,lubana2022orchestra,han2022fedx,zhuang2022divergence} & $\times$ & CV & 1 & 60000 & Dirichlet \& Uniform\\
        CIFAR-100 & \cmark \cite{liang2022rscfed,diao2022semifl} & \cmark \cite{lubana2022orchestra,zhuang2021collaborative,zhuang2022divergence} & $\times$ & CV & 1 & 60000 & Dirichlet \& Uniform\\
        SVHN & \cmark \cite{wang2022enhancing,diao2022semifl} & \cmark \cite{han2022fedx} & $\times$ & CV & 1 & 73257 & Dirichlet \& Uniform\\
        Sent140 & \cmark \cite{fan2022private} & $\times$ & $\times$ & NLP & 1 & 1600498 & Natural (Twitter User)\\
        Reuters & \cmark \cite{itahara2021distillation} & $\times$ & $\times$ & NLP & 1 & 11228 & Dirichlet\\
        IMDb & \cmark \cite{itahara2021distillation} & $\times$ & $\times$ & NLP & 1 & 50000 & Dirichlet\\
        Landmark-23K & $\times$ & \cmark \cite{he2021ssfl} & $\times$ & CV & 1 & 1600000 & Natural (Location)\\
        Digit-Five & $\times$ & $\times$ & \cmark \cite{peng2020federated,gan2022fruda} & CV & 5 & 107348 & Natural (Style)\\
        Office-Caltech10 & $\times$ & $\times$ & \cmark \cite{peng2020federated,gan2022fruda,kang2022communicational} & CV & 4 & 2533 & Natural (Style)\\
        DomainNet & $\times$ & $\times$ & \cmark \cite{peng2020federated,gan2022fruda} & CV & 6 & 416401 & Natural (Style)\\
        AmazonReview & $\times$ & $\times$ & \cmark \cite{peng2020federated} & NLP & 4 & 8000 & Natural (Product Category)\\
        Mic2Mic & $\times$ & $\times$ & \cmark \cite{gan2022fruda} & Speech & 4 & 65000 & Natural (Device Type)\\
        GTA5 & $\times$ & $\times$ & \cmark \cite{yao2022federated} & CV & 4 & 25000 & Natural (Location)\\
        
    \bottomrule
    \end{tabular}}
    \caption{Commonly used datasets for evaluating FL methods without full labels. \cmark and $\times$ indicate that the dataset has or has not been used for evaluating an FL setting without full labels, respectively. \# domains, \# samples denote the number of domains and the total number of samples in the dataset. Datasets with multiple domains are more commonly used for unsupervised federated transfer learning. }
    \label{tab:flud_dataset}
\end{table}
Benchmarking datasets are important for the development of machine learning research. In this section, we introduce commonly used datasets and benchmarks for the problem of FL without full labels in the existing literature. A summary of datasets can be found in Table \ref{tab:flud_dataset}. We find out that for both federated semi-supervised and unsupervised learning, existing works mainly partition (e.g. according to Dirichlet distributions) datasets for centralized machine learning (e.g. CIFAR-10, CIFAR-100, SVHN) manually, and manually sample a subset of labels. On the contrary, for federated transfer learning, datasets generally form natural partitions (e.g. city in GTA5, product types in AmazonReview, etc.) based on different domains. We thus conclude that real-world datasets representing realistic data heterogeneity and label isolation problems are still needed to credibly evaluate federated semi-supervised and self-supervised methods. 

\section{Related Surveys}\label{sec:related}
Federated learning has attracted the attention of researchers worldwide. Therefore, there have been many survey papers that cover various aspects of FL. In this section, we summarize and analyze existing survey papers compared to our work. Table \ref{tab:related} shows a summary of comparisons between related surveys and ours. 

First, our work differs from general surveys on FL \cite{kairouz2021advances,yang2019federated,li2020federatedsurvey} in that they provide comprehensive reviews on a wide range of FL aspects, including privacy preservation, communication reduction, straggler mitigation, incentive mechanisms, etc. Among them, communication and privacy are also important issues in the problem of FL without full labels and are covered in our survey. On the contrary, our survey is focused on a specific aspect, namely how to deal with unlabeled data. Second, our work also differs from surveys on semi-supervised learning \cite{chapelle2009semi}, self-supervised learning \cite{liu2021self}, and transfer learning \cite{pan2010survey} in the centralized setting, in that while they extensively summarize machine learning techniques for unlabeled data, they fail to cover FL-specific challenges, such as label isolation, data privacy, etc. Finally, compared to surveys that focus on FL algorithms on non-i.i.d. data \cite{wang2021field,li2022federated,zhu2021federated}, our work focuses on leveraging unlabeled data to assist FL, while these surveys focus on FL with fully labeled data, but are not independent and identically distributed. Nonetheless, these surveys are related to our work in that non-i.i.d. data is an important challenge in all FL settings, and we also summarize how existing works address the challenge in the problem of FL without full labels. 

The most related survey to our work is \cite{gao2022survey}, which surveyed FL techniques to tackle data space, statistical, and system heterogeneity. Our work is similar to \cite{gao2022survey} in two ways. On one hand, statistical heterogeneity is a key challenge in FL, and we also summarize how existing works address the challenge in FL without full labels. On the other hand, homogeneous and heterogeneous FTL (Section \ref{sec:fed-utl}) are powerful tools to solve statistical and data space heterogeneity, respectively, which are also covered in Sections 3 and 4 in \cite{gao2022survey}. Nonetheless, the main focus of \cite{gao2022survey} lies in supervised FL with labeled data, which is different from our work which additionally covers federated semi-supervised and self-supervised methods.

\section{Conclusion and Future Directions}\label{sec:conclu-future}
\begin{table}[]
    \centering
    \resizebox{\linewidth}{!}{
    \begin{tabular}{ccc}
    \toprule
        \textbf{Survey Papers} & \textbf{Similarities} & \textbf{Differences}\\
        \midrule
        \multirowcell{4}{\cite{yang2019federated,kairouz2021advances,li2020federatedsurvey}} & \multirowcell{4}{Similar to our survey, these papers \\review existing solutions to protect \\data privacy and reduce \\communication/computation overhead.} & \multirowcell{4}{These papers cover a wide range of \\aspects in general FL, while \\our survey focuses on a specific problem \\of leveraging unlabeled data. }\\
        & \\
        & \\
        & \\
        \midrule
        \multirowcell{4}{\cite{pan2010survey,jing2020self,liu2021self,chapelle2009semi,van2020survey,zhuang2020comprehensive}} & \multirowcell{4}{Similar to our survey, these papers\\review machine learning methods \\for unlabeled data, including semi-supervised, \\self-supervised, and transfer learning.} & \multirowcell{4}{These papers do not cover \\FL specific challenges, such as \\labeled data isolation, data heterogeneity, \\data privacy, etc.  }\\
        & \\
        & \\ 
        & \\
        \midrule
        \multirowcell{4}{\cite{li2022federated,wang2021field,zhu2021federated}} & \multirowcell{4}{Similar to our survey, these papers \\review methods in FL that \\address the problem of non-i.i.d. data\\ (i.e. data heterogeneity). } & \multirowcell{4}{These papers primarily focus on\\ optimization algorithms for fully supervised FL,\\ while our work focuses specifically\\ on leveraging unlabeled data. }\\
        & \\
        & \\
        & \\
        \midrule
        \multirowcell{4}{\cite{gao2022survey}} & \multirowcell{4}{Similar to our survey, \cite{gao2022survey} covers\\methods to tackle data heterogeneity.\\Also, \cite{gao2022survey} reviews existing works\\ on homogeneous and heterogeneous FTL. } & \multirowcell{4}{\cite{gao2022survey} primarily focuses on heterogeneity\\ in supervised FL, while our work focuses on\\ leveraging unlabeled data and covers \\federated semi-supervised and self-supervised learning.  }\\
        & \\
        & \\
        & \\
    \bottomrule
    \end{tabular}}
    \caption{Comparative analysis between our survey and related surveys. }
    \label{tab:related}
\end{table}

\begin{table}[]
    \centering
    \resizebox{\linewidth}{!}{
    \begin{tabular}{cccc}
        \toprule
        \textbf{Learning Paradigm} & \textbf{Main Techniques} & \textbf{Advantages} & \textbf{Disadvantages}  \\
        \midrule
        \multirowcell{3}{Federated Semi-\\supervised Learning} & \multirowcell{9}[0pt][l]{Enhancing methods in centralized settings with\\
         1. \textbf{Label isolation}: Pseudo-labeling, \\domain alignment, etc. \\
        2. \textbf{Privacy}: DP, HE, etc. \\
        3. \textbf{Data heterogeneity}: Source domain selection, \\divergence-aware update, etc. \\
        4. \textbf{Efficiency tradeoff}: Sample selection, \\communication reduction, HE optimization, etc. } &
        \multirowcell{3}{Similar formulation to conventional FL.\\Can directly perform supervised tasks.} & \multirowcell{3}{Data heterogeneity inherently violates\\i.i.d. assumption. Large-scale unlabeled data \\creates an efficiency tradeoff. } \\
        & \\
        & \\
        \cmidrule{1-1} \cmidrule{3-4}
        \multirowcell{3}{Federated Self-\\supervised Learning} & & \multirowcell{3}{Full utilization of client data. \\Suitable for unsupervised tasks \\like retrieval, clustering, etc. } & \multirowcell{3}{Data heterogeneity inherently violates\\i.i.d. assumption. Need labels for \\supervised tasks. }\\
        & \\
        & \\
        \cmidrule{1-1} \cmidrule{3-4}
        \multirowcell{3}{Federated \\Transfer Learning} & &\multirowcell{3}{Models data heterogeneity, which is \\a key challenge in FL. Flexible\\ formulation (heterogeneous FTL). } & \multirowcell{3}{Source domain selection requires \\intricate design or manual effort.} \\
        & \\
        & \\
         \bottomrule
    \end{tabular}}
    \caption{A summary of techniques, advantages, and disadvantages of learning paradigms reviewed in this paper.  }
    \label{tab:summary}
\end{table}
\subsection{Summary of the Survey}
In this paper, we present a survey about the problem of federated learning without full labels. We introduce three learning paradigms to solve the problem, federated semi-supervised learning, federated self-supervised learning, and federated transfer learning. We further review existing works in these paradigms and discuss how they address the crucial challenges, i.e. label isolation, privacy protection, data heterogeneity, and efficiency tradeoff. Table \ref{tab:summary} shows a summary of the main techniques, advantages, and disadvantages of learning paradigms discussed in this paper. We finally present a summary of the datasets and benchmarks used to evaluate FL methods without full labels. 

\subsection{Future Directions}
Compared to general FL with full supervision, the problem of FL without full labels is still under-explored. We highlight the following future directions in the context of FL without full labels. 

\subsubsection{Trustworthiness}
Trustworthiness is an important aspect in real-world machine learning systems like FL. Generally speaking, users of machine learning systems would expect a system to be private, secure, robust, fair, and interpretable, which is what trustworthiness mean in the context of FL. 

Unlabeled data can play an important role in enhancing trustworthiness from multiple aspects. 
\begin{itemize}
    \item \textbf{Robustness}: A robust system requires that its output should be insensitive to small noises added to the input. A machine learning system that is not robust can significantly compromise its security in real-world applications. For example, studies \cite{kurakin2018adversarial} have shown that is it possible to tweak physical objects to fool an object detection model. In applications like autonomous driving, this property becomes a security threat. 

    Many research works have studied how to enhance robustness with unlabeled data \cite{deng2021improving,carmon2019unlabeled}. For example, Carmon et al. and Uesato et al. \cite{carmon2019unlabeled,alayrac2019labels} show that pseudo-labeling, one of the most common semi-supervised learning techniques, can boost the robustness by 3-5\% over state-of-the-art defense models. Deng et al. \cite{deng2021improving} additionally find out that even out-of-distribution unlabeled data helps enhance robustness. Therefore, how these techniques can be adapted in the FL setting with heterogeneous data is an interesting future direction. Also, as common methods of learning robust models (i.e. adversarial training \cite{madry2018towards}) are inefficient, it is promising to study whether FL methods without full labels can be an efficient substitute. 

    \item \textbf{Privacy}: In real-world machine learning applications, labeling data itself is a compromise of data privacy, as domain experts have to directly observe the data. Therefore, solving the FL problem without full labels inherently leads to better data privacy. In addition, unlabeled data provides a better way of navigating through the privacy-utility tradeoff in differential privacy (DP) \cite{dwork2014algorithmic}. For example, PATE \cite{papernot2017semi} shows that with an additional set of unlabeled data, it simultaneously achieves a higher model accuracy and a tighter privacy bound compared to the state-of-the-art DPSGD method \cite{abadi2016deep}. Therefore, how to select and leverage unlabeled data to aggregate client knowledge privately while maintaining good model accuracy is also a promising direction. 

    \item \textbf{Interpretability}: Interpretability indicates that a machine learning system should be able to make sense of its decision, which generally creates trust between users and system developers. There are many ways to instill interpretability in machine learning, among which disentangled representation learning \cite{lee2018diverse} is a popular direction. Informally speaking, disentangled representation aims to map the inputs to latent representations where high-level factors in the input data are organized in a structured manner in the representations (e.g. brightness, human pose, facial expressions, etc.). Thus, disentangled representations provide intuitive ways to manipulate and understand deep learning models and features. 

    Much progress has been made in unsupervised disentangled representation learning. For example, InfoGAN \cite{chen2016infogan} learns disentangled representations by maximizing the mutual information between the features and the output. Beta-VAE \cite{higgins2017beta} disentangles features by adding an independence regularization on the feature groups. Therefore, it is promising to instill interpretability in FL via unlabeled data with disentangled representations. In FL, the participants commonly hold data with varying data distributions. Therefore, how to stably disentangle the heterogeneous feature distributions from multiple participants is a challenge for interpretable FL without full labels. 

    \item \textbf{Fairness}: As machine learning models are increasingly involved in decision-making in the daily lives of people, the models should not discriminate one group of users against another (e.g. gender, race, etc.). Informally speaking, the fairness of a machine learning model $g_\theta$ over a sensitive attribute $s$ can be described as the difference between the model performances given different values of $s$, 
    \begin{equation}
        \Delta_{s, \theta} = \|m(g_\theta|s=1) - m(g_\theta|s=0)\|,
        \label{eqn:fairness}
    \end{equation}
    where $\|\cdot \|$ is a distance, and $m(g_\theta|s=1)$ is a performance metric stating how well the model performs when the sensitive attribute $s=1$. 
    
    When $m(g_\theta|s)$ does not involves labels (e.g. some groups have a higher probability to be predicted positive), FADE \cite{hong2021federated} provides a good solution to ensure group fairness. However, when $m(g_\theta|s=1)$ requires labeled data (e.g. classification accuracy is lower for under-represented groups), enforcing fairness with unlabeled data remains an open problem, both for general machine learning and FL. 
\end{itemize}

\subsubsection{Generalization to Unseen Domains}
All the introduced techniques in this paper require at least observing the test domain such that it can work well on it. Even for federated transfer learning, some unlabeled samples in the target domain are still needed for successful adaptation. However, in real-world applications, it is often required to adapt to completely unseen domains. For example, FL models should try to adapt to new users that constantly join mobile applications, who, at the time of joining, have no interaction data available. The problem setting triggers research in federated domain generalization (FedDG). However, existing works in FedDG \cite{liu2021feddg,nguyen2022fedsr} assume that all domains are fully labeled, which, as stated in this survey, is not realistic. It is thus important to study the FedDG problem under limited labeled data and large-scale unlabeled data. 

\subsubsection{Automatic FL without Full Labels}
Automatic machine learning (AutoML) \cite{he2021automl} is a class of methods that aim to achieve good model performances without manual tuning (e.g. architecture, hyperparameters, etc.). In FL without full labels, as different participants may hold heterogeneous labeled or unlabeled data, it may not be optimal for them to share the same model architecture. Integrating AutoML to FL without full labels thus enables participants to find personalized architectures to achieve the performance-efficiency tradeoff. However, participants with only unlabeled data cannot independently evaluate the performance of the model, creating challenges to automatic FL without full labels.

\end{document}